# Novel and Tuneable Method for Skin Detection Based on Hybrid Color Space and Color Statistical Features

Reza Azad*, Hamid Reza Shayegh Brojeeni**
*IEEE Member, Shahid Rajaee Teacher Training University, Tehran, Iran, rezazad68@gmail.com
**Assistant Professor, Shahid Rajaee Teacher Training University, Tehran, Iran, hamid.Shayegh@gmail.com

*Abstract:* *Skin detection is one of the most important and primary stages in some of image processing applications such as face detection and human tracking. So far, many approaches are proposed to done this case. Near all of these methods have tried to find best match intensity distribution with skin pixels based on popular color spaces such as RGB, CMYK or YCbCr. Results show these methods cannot provide an accurate approach for every kinds of skin. In this paper, an approach is proposed to solve this problem using statistical features technique. This approach is including two stages. In the first one, from pure skin statistical features were extracted and at the second stage, the skin pixels are detected using HSV and YCbCr color spaces. In the result part, the proposed approach is applied on FEI database and the accuracy rate reached 99.25 ± 0.2. Further proposed method is applied on complex background database and accuracy rate obtained 95.40±0.31%. The proposed approach can be used for all kinds of skin using train stage which is the main advantages of it. Low noise sensitivity and low computational complexity are some of other advantages.*

**1. Introduction**

Image processing now is one of the most scientific courses, which can use in very applicable and industrial projects. Some of these popular projects are face detection [1], human tracking [2, 3], Human identification, visual tracking for surveillance, hand gesture recognition, searching, image retrieval and filtering image contents on the web and many others. One the most stages which should be used in these cases, is skin detection. In this respect, many approaches have proposed to detect skin which provide high detection rate. Some of them are used RGB color space to solve this problem such as [4,5] and [6]. Some of the researchers used other color spaces such as YCbCr [7, 8], HSV [9], CIE LUV [10], and Farnsworth UCS [11]. As a common algorithm, near all of them tried to find the channel intensities which are too match by skin pixels in images. Also, in some methods the researchers used texture analysis approaches. Wu et al. used wavelet filters to done it accurately [11]. There are very various kinds of skins such as white, black, red, approximately green and etc. one of the most important mentions in this case is to detect all kinds of skins accurately. Also insensitivity to illumination and noise are some other problems in this case. According to this mentions, in this paper an approach is proposed based on hybrid color space and color statistical technique. In this approach which contains two stages, first of all, some train images which include just pure skin pixels of slightly kind of skin is provided. Next, mean and variance of these pixels are computed for each HSV and YCbCr channels individually. Then adaptable threshold value will be obtained by this feature and finally the threshold will be applied on the test image to skin detection. In the result part, proposed approach is applied on the FEI and complex background database images and the accuracy rate is computed. Some of the proposed approach advantages are: a) Adaptability to most kinds of skin by using a train stage; b) Low complexity in computation and time; c) Low sensitivity to illumination by using adaptable threshold. In the next section we have explained the Proposed method for skin detection. Experimental results have been explained in third section and fourth section is conclusion.

**2. Proposed Method**

Proposed method, consisted of 2 stages which in the first stage, some train images which contains just pure skin pixels, should be provided. Then from pure skin statistical features are extracted. In the Stage1 for each group of skin such as white, black, red, approximately green and etc. skin feature just one time will be calculated. And at the second stage, the skin pixels are detected with use of HSV and YCbCr color spaces. General diagram of first stage is shown in Fig. 1.

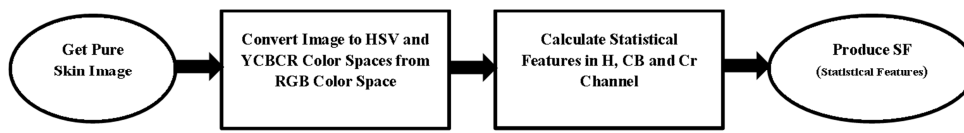

Fig. 1: Stage 1 pure skin feature extraction

## 2.1 Extracting Statistical Feature from Pure Skin

The skin feature in color image must be calculated. In order to achieve this, skin color space needs to be defined. As HSV and YCbCr color space has resemblance with human vision, we convert RGB color space into HSV and YCbCr color space. Using Equation (1), RGB color space is converted to HSV color space [12].

$$H = a\,Cos(0.5 * \frac{2*R - G - B}{\sqrt{(R-G)^2 + (R-B)^2}})) * 57.32$$

$S = (Max(R, G, B) - Min(R, G, B))/Max(R, G, B)$

$V = Max(R, G, B)/255.0$ 

(1)

The equivalent YCbCr matrix for RGB image [13] is given by Equation (2).

$$\begin{bmatrix} Y \\ Cb \\ Cr \end{bmatrix} = \begin{bmatrix} 0.2568 & 0.5041 & 0.0980 \\ -0.1482 & -0.2910 & 0.4392 \\ 0.4392 & -0.3678 & -0.0714 \end{bmatrix} \begin{bmatrix} R \\ G \\ B \end{bmatrix} + \begin{bmatrix} 16 \\ 128 \\ 128 \end{bmatrix} \quad (2)$$

In order to extract the skin feature, it's enough to compute statistical features like mean and variance of skin pixels in each channel. It is shown in Equation (3) and Equation (4) for H channel.

$$\text{Mean}(H) = 1/{n \times m} \sum_{1 \leq i \leq n} \sum_{1 \leq j \leq m} H(i,j) \quad (3)$$

$$Var(H) = 1/{n \times m} \sum_{\substack{1 \leq i \leq m \\ 1 < j < n}} (H(i,j) - mean(H))^2 \quad (4)$$

Where, $H(i,j)$ the hue value of pixel in $i_{th}$ row and $j_{th}$ column in H channel. Also, n and m are the size of image. By using a similar way, the mean and Variance in CB and CR channels can be computed. Now, the statistical feature like SF which is shown in Equation (5) can be provided as a good identification of skin pixels in each H, CB and CR channel.

$$SF = (Mean_H, Mean_{CB}, Mean_{CR}, Var_H, Var_{CB}, Var_{CR}) \quad (5)$$

## 2.2 Skin detection

In this stage skin region will be extracted. In order to achieve this region, at first, entrance image is converted to the HSV and YCbCr color space, then skin region will be extracted by use of SF. General diagram of the second stage is shown in Fig. 2.

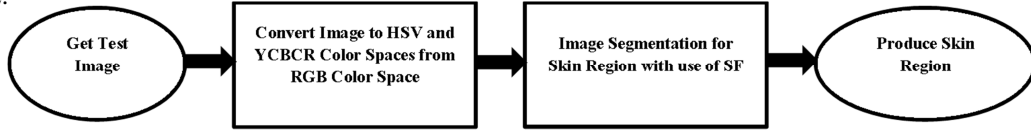

Fig. 2: Skin detection on test image

## 3.1 Skin Segmentation Based on HSV and YCbCr color space

A frontal image from FEI face database is considered for processing. The original image is as shown in Fig. 3(a). The following function is used for skin segmentation.

---

Function Skin Segmentation (Orginal.image, SF)

1) $(\forall I, J: I \in Orginal.image(rows)\ AND\ J \in Orginal.image(columns))\ Do$

2) IF $(Mean_H - Var_H \leq H(I,J) \leq Mean_H + Var_H)$ is TRUE AND

3) $(Mean_{Cb} - Var_{Cb} \leq Cb(I,J) \leq Mean_{Cb} + Var_{Cb})$ is TRUE AND

4) $(Mean_{Cr} - Var_{Cr} \leq H(I,J) \leq Mean_{Cr} + Var_{Cr})$ is TRUE THEN

5) Denote Orginal.Image (I, J) Pixel as skin Pixel; Write 1 ELSE

6) Orginal.Image (I, J) Pixel as Non-skin Pixel; Write 0

End Function

---

Where, I and J are respective row and column coordinates of original RGB image denoted by 'Original. Image'. As per the given pseudo code, if pixel satisfies all conditions imposed on Cr, Cb and Hue, it is affirmed as skin pixel, else it is non-skin pixel. As we wrote '1' and '0' for skin and non-skin pixels respectively, we get Fig. 3(b) as binary image.

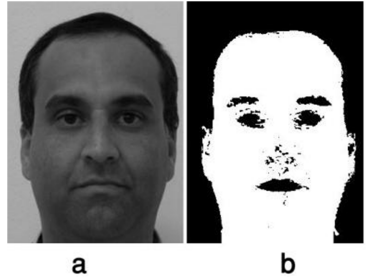

Fig. 3: (a) Original Image   (b) skin detected using proposed approach

### 3. Result and Discussion

Our suggestive method have been done on Intel Core i3-2330M CPU, 2.20 GHz with 2 GB RAM under Matlab environment. Fig. 4 shows the face of worked systems. In this study, for experimental analysis, we considered a FEI and complex background database. The FEI face database is a Brazilian face database that contains a set of images taken between June 2005 and March 2006 at the Artificial Intelligence Laboratory of FEI in São Bernardo do Campo, São Paulo, Brazil. The results are as presented in table I.

TABLE I: Detection Rate on FEI Database

| Number of Images | FEI color image database | |
|---|---|---|
| | Successful Skin Detection | Detection Rate |
| 400 | 400 | 99.25 ± 0.2 |

The detection rate that we used for evaluation our method is showing in Equation (6) that mentioned in [5].

$$DetectionRate = 100 \times \left(\frac{N_{DD} + N_{SS}}{m * n}\right) \quad (6)$$

Where, n and m are the size of test image. N_SS means the number of really skin pixels which detected as skin and N_DD means the number of pixels which are not skin and the proposed approach don't detects them as skin. The average of accuracy rates for all of the test images in FEI database is computed 99.25 ± 0.2.

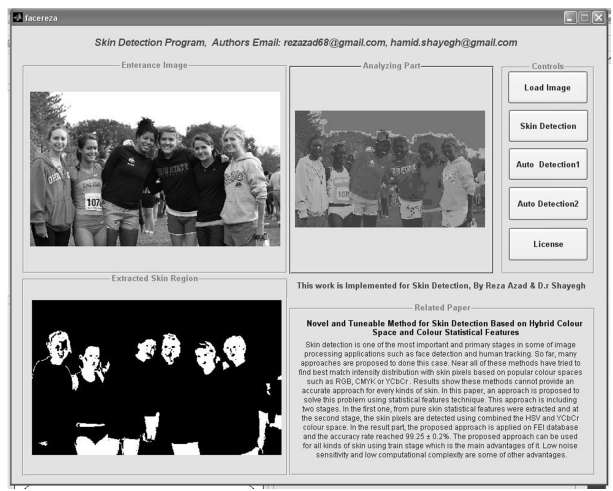

Fig. 4: Skin detection system

High detection rate shows the quality of proposed approach to use in every applications, which are needed a skin detection stage. Low complexity in computation and time are some of other advantages of the proposed approach. Using a train stage is provide, various skin kinds detection ability. Also proposed approach can be used for skin kind's classification in complex images also. Complex images mean the images which are included two or more kinds of skins. Further we tasted our method on complex background images for showing high accuracy of our method. The results are as presented in table 2. Fig.5 shows the sample of these images.

TABLE 2: Detection Rate on Complex Background Database

| Number of Images | Complex Background images Database | |
|---|---|---|
| | Successful Skin Detection | Detection Rate |
| 100 | 100 | 95.40 ± 0.31 |

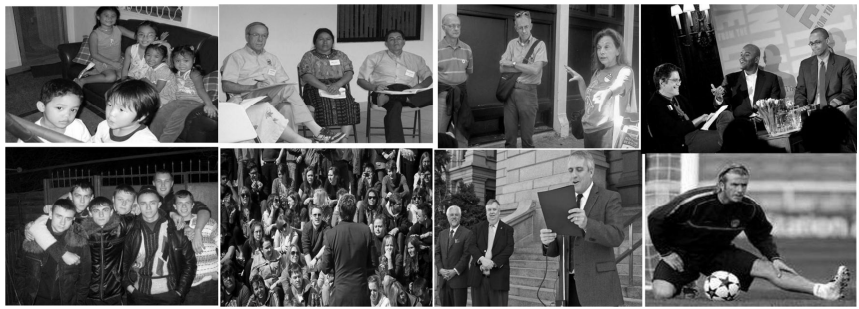

Fig. 5: Some of images with complex background that system recognized them correctly

## 4. Conclusion

The aim of this paper was to propose an accurate approach to detect skin in facial or else images. According to this aim, an approach is proposed based on statistical feature techniques and using a train stage. In the train stage, mean and variance of the pure skin pixels are computed in each H, Cb and Cr channel individually. Next, by using this features that can provide a good severability between skin and non-skin pixels, skin region in hybrid color space were extracted. The result part is proved the quality of proposed approach to detect skin pixels in terms of skin kinds. In this respect, the accuracy may decrease when the background pixels have more similarity to skin pixels which can be solved using statistical features. We tested our method on FEI and complex background database and the accuracy rate reached 99.25 ± 0.2 and 95.40 ± 0.31 respectively. The proposed approach is a multipurpose, which can be used for skin detection and skin kind classification problems. The proposed approach can be used for all skin kinds against many previous ones which is the main advantage of this paper. Low computational complexity and low sensitivity to noise are some others.